\documentclass[letterpaper]{article} 
\usepackage{aaai24}  
\usepackage{times}  
\usepackage{helvet}  
\usepackage{courier}  
\usepackage[hyphens]{url}  
\usepackage{graphicx} 
\usepackage{natbib}  
\usepackage{caption} 
\usepackage{algorithm}
\usepackage{algorithmic}

\usepackage{newfloat}
\usepackage{listings}
\usepackage{color}

\usepackage{amsmath}
\usepackage{amssymb}
\usepackage{mathtools}
\usepackage{amsthm}

\usepackage{times}
\usepackage{epsfig}
\usepackage{graphicx}
\usepackage{multirow}
\newcommand{\ygy}[1]{#1}

\def\ie{\emph{i.e.,~}}

\def\etc{{\em etc}}

\DeclareMathOperator*{\onehot}{OneHot}

\DeclareMathOperator*{\upsample}{Upsample}
\DeclareMathOperator*{\flatten}{Flatten}
\DeclareMathOperator*{\argmin}{arg\,min}

\newcommand{\abs}[1]{\left|#1\right|}
\usepackage{bbding}
\usepackage{amssymb}
\usepackage{pifont}
\newcommand{\cmark}{\ding{51}}
\newcommand{\xmark}{\ding{55}}
\newcommand{\AP}{\text{AP}}

\DeclareCaptionStyle{ruled}{labelfont=normalfont,labelsep=colon,strut=off} 
\floatstyle{ruled}
\newfloat{listing}{tb}{lst}{}
\floatname{listing}{Listing}
\title{Semantic-Aware Transformation-Invariant RoI Align}
\author{
    Guo-Ye Yang\textsuperscript{\rm 1},
    George Kiyohiro Nakayama\textsuperscript{\rm 2},
    Zi-Kai Xiao\textsuperscript{\rm 1},
    Tai-Jiang Mu\textsuperscript{\rm 1}\thanks{Corresponding author is Tai-Jiang Mu.},
    Xiaolei Huang\textsuperscript{\rm 3},
    Shi-Min Hu\textsuperscript{\rm 1}
}
\affiliations{
    \textsuperscript{\rm 1}Department of Computer Science and Technology, BNRist, Tsinghua University\\
    \textsuperscript{\rm 2}Stanford University\\
    \textsuperscript{\rm 3}College of Information Sciences and Technology, Pennsylvania State University\\
    yanggy19@mails.tsinghua.edu.cn, w4756677@stanford.edu, xzk23@mails.tsinghua.edu.cn, \\
    taijiang@tsinghua.edu.cn, suh972@psu.edu,  shimin@tsinghua.edu.cn

}

\usepackage{bibentry}

\begin{document}

\maketitle


\begin{abstract}
Great progress has been made in learning-based object detection methods in the last decade.
Two-stage detectors often have higher detection accuracy than one-stage detectors, due to the use of region of interest (RoI) feature extractors which extract transformation-invariant RoI features for different RoI proposals, making refinement of bounding boxes and prediction of object categories more robust and accurate. 
However, previous RoI feature extractors can only extract invariant features under limited transformations.
In this paper, we propose a novel RoI feature extractor, termed Semantic RoI Align (SRA),
which is capable of extracting invariant RoI features under a variety of transformations for two-stage detectors.
Specifically, we propose a semantic attention module to adaptively determine different sampling areas by leveraging the global and local semantic relationship within the RoI.
We also propose a Dynamic Feature Sampler which dynamically samples features based on the RoI aspect ratio to enhance the efficiency of SRA,
and a new position embedding, \ie Area Embedding, to provide more accurate position information for SRA through an improved sampling area representation.
Experiments show that our model significantly outperforms baseline models with slight computational overhead.
In addition, it shows excellent generalization ability and can be used to improve performance with various state-of-the-art backbones and detection methods.
\end{abstract}
\section{Introduction}
  
\begin{figure}[!t]
	\centering
	\includegraphics[width=\linewidth]{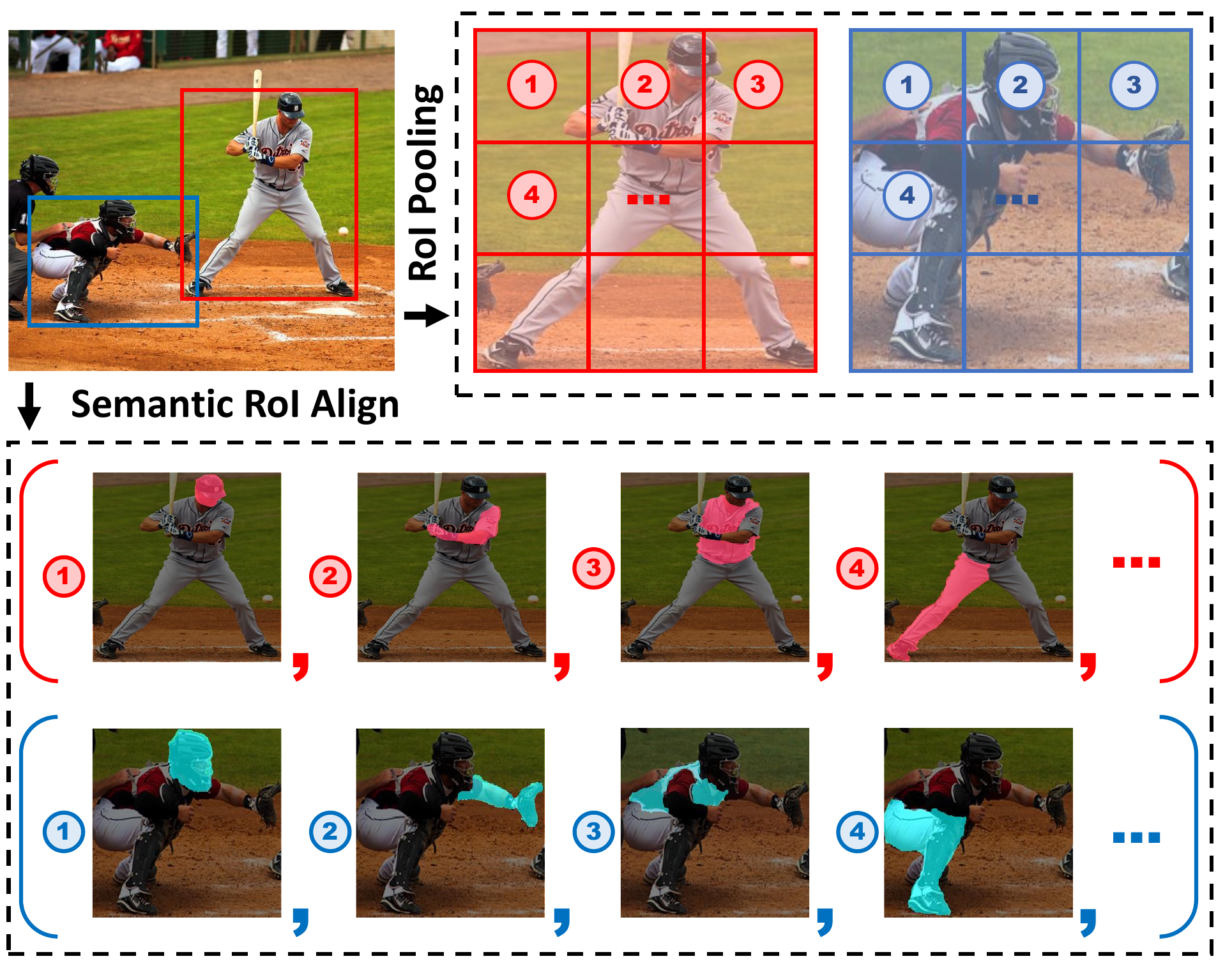}
	\caption{Previous RoI feature extractor versus the proposed Semantic RoI Align (SRA). Top: RoI Pooling samples each feature in some specific positions, making extracted RoI features sensitive to object poses. Bottom: SRA samples each feature from different semantic regions, making it capable of extracting invariant RoI features under various transformations including object pose transformation.}
	\label{proposed}
\end{figure}
  
As a fundamental computer vision task, object detection aims to locate and recognize objects of interest in input images.
In the last decade, great progress has been made in learning-based object detection methods, making them widely useful in our daily uses, such as face recognition, text detection, pedestrian detection, among others.

Most existing detection methods can be grouped into two categories, \ie one-stage detectors~\cite{liu2016ssd,redmon2016you} and two-stage detectors~\cite{ren2015faster,he2017mask}. 
One-stage detectors directly predict objects with a single neural network in an end-to-end manner. In contrast, two-stage detectors first propose a list of object proposals and then predict the proposals' labels and refine bounding boxes with extracted RoI features of each proposal. 
RoI features provide better transformation-invariance for different proposal regions and using them 
can thus better refine bounding boxes and predict the category of each proposal in the second stage of a two-stage detector. In this paper, we mainly focus on improving the RoI feature extractor for two-stage detectors.

Similar objects may show great appearance differences in images due to different environmental conditions, object poses, \etc., making detection difficult to be generalizable under various transformations~\cite{girshick2015fast}.
Therefore, many RoI feature extractors aim to extract transformation-invariant object features.
RoI Pooling~\cite{girshick2015fast} is the pioneering work for RoI feature extraction, which pools features in a fixed number of sub-regions of the RoI and obtains scale-invariant features.
RoI Align~\cite{he2017mask} further improves the positional accuracy of RoI Pooing via bilinear interpolation.
RoI Transformer~\cite{ding2019learning} extracts rotation-invariant features by rotating sampling positions with a regressive rotation angle.
However, previous methods cannot extract invariant features
for more complex transformations like perspective transformation and object pose transformation. 
Though there exist works like Deformable RoI Pooling (DRoIPooling) ~\cite{dai2017deformable, zhu2019deformable} that can extract invariant features under some complex transformations 
by adaptively adding a regressive offset to each sampling position, experiments show that it only achieves invariance under scaling transformation and cannot easily be extended to handle others such as rotation.
This is because the sampling position offsets are regressed with convolutional networks, which need to be trained with transformed data~\cite{krizhevsky2012imagenet} and different transformations would require different kernels since the same position of a convolutional kernel may correspond to different object regions when the object is transformed under different transformations.

In this paper, we regard different transformations like perspective and pose transformations as being comprised of spatial transformations of different semantic parts, 
while also considering that high-level features of semantic regions are more stable under varying transformations.
From this perspective, we propose a Semantic RoI Align (SRA) to extract transformation-invariant RoI features by sampling features from different semantic regions.
RoI Pooling samples features at specific locations in a RoI. A limitation of such sampling can be seen from the example shown in Figure~\ref{proposed} (top row); 
the 4-th sampling location extracts features of the background in the red RoI, whereas that location extracts features of a player's leg in the blue RoI. Such sampling loses invariance under pose transformations. 
In contrast, in our proposed SRA, we design a \ygy{semantic attention module} to obtain different semantic regions by leveraging the global and local semantic relationship within the RoI. Then we sample features from the semantic regions, and concatenate the sampled features as the RoI feature which is semantic-aware and transformation-invariant. 

Since the computational efficiency of SRA determined by the feature sampling resolution, we propose a Dynamic Feature Sampler to dynamically sample features according to the aspect ratio of different RoIs, which speeds up SRA while minimizing the impact on accuracy.
Furthermore, previous positional embedding methods~\cite{zhao2020exploring} only encode information of the sampling center, which cannot accurately represent regional information. We thus propose a new positional embedding, namely Area Embedding, which embeds positions in a sampling area into a fixed-length vector, providing more accurate position information.
SRA can replace the RoI feature extractor in most two-stage detectors and brings higher detection accuracy with a slight overhead in the number of network parameters and computation.

By using SRA as RoI feature extractor for Faster R-CNN~\cite{ren2015faster},
our method achieves 1.7\% higher mAP in the COCO object detection task with only additional 0.2M  parameters and 1.1\% FLOPs compared to the baseline model.
Meanwhile, it also exceeds other RoI feature extractors with less computational overhead.
To verify the generalizability of SRA, 
we equip it to various state-of-the-art backbones and detection methods. Results show that SRA can consistently boost their detection accuracy.

In summary, our contributions are:

\begin{itemize}
\item a novel RoI feature extractor, \ie Semantic RoI Align, which is able to extract transformation-invariant RoI features and can be plugged into most two-stage detectors to improve detection accuracy with little extra cost,

\item a Dynamic Feature Sampler which makes SRA implementation efficient, and an Area Embedding which provides more comprehensive and accurate information of sampled positions.

\item Extensive experiments that demonstrate the superiority of the SRA and its great generalizability to various state-of-the-art backbones and detection methods.
\end{itemize}

\section{Related Work}
\subsection{Object detection and RoI feature extractors}
In recent years, deep learning techniques are dominant in object detection.
Most deep object detection methods can be categorized into two types: one-stage detectors~\cite{redmon2016you} and two-stage detectors~\cite{girshick2015fast}.
Faster R-CNN~\cite{ren2015faster} is a two-stage network with a Regional Proposal Network (RPN) predicting multiple RoI proposals; RoI features are then extracted by an RoI feature extractor to 
predict object bounding boxes and categories in the second-stage network. 
Mask R-CNN~\cite{he2017mask} proposed a general framework for object instance segmentation tasks.
Dynamic head~\cite{dai2021dynamic} proposed to use scale, spatial, and task-aware attention mechanisms to improve detection accuracy.

\begin{figure*}[ht]
	\centering
	\includegraphics[width=1\linewidth]{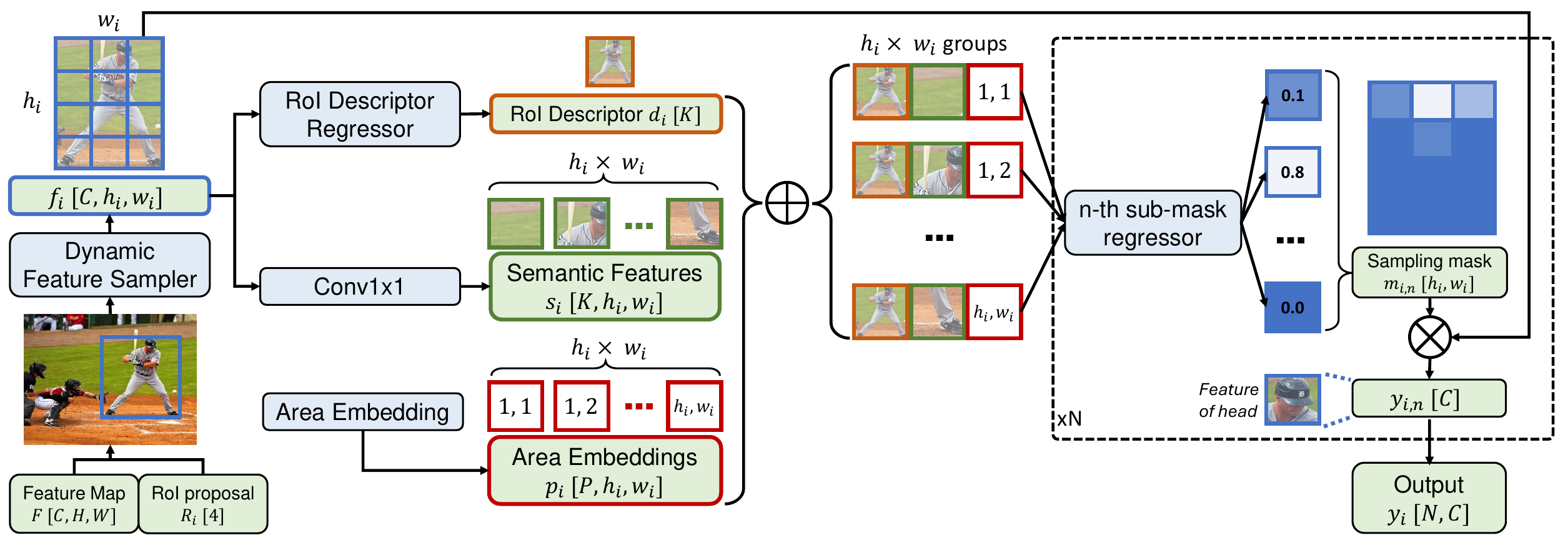}
	\caption{\ygy{The network architecture of our Semantic RoI Align.  $\bigotimes$ means matrix multiplication, and $\bigoplus$ means  concatenation.}}
	\label{URA}
\end{figure*}

RoI feature extractors are used to extract transform-invariant features in two-stage detectors, so that the second-stage network can refine the bounding boxes and predict object categories more accurately.
RoI Pooling~\cite{girshick2015fast} performs scale-invariant feature extraction by dividing the RoI into a fixed number of bins, pooling the features in each bin, and concatenating them into a vector of fixed size. 
RoI Align~\cite{he2017mask} uses bilinear interpolation to more accurately extract features. 
RoI Transformer~\cite{ding2019learning} extracts rotation-invariant features by correcting the extracted features using a learned rigid transformation supervised by ground-truth oriented bounding boxes.
However, these methods only model features invariant to rigid transformations, while ignoring non-rigid transformations. 
Deformable RoI Pooling~\cite{dai2017deformable, zhu2019deformable} extracts features by adding a regressive offset to each sampling position of RoI Pooling.
Our experiments show that it can only extract invariant features under scale transformation which could be due to its learning the regressive offset by convolution, making it hard to generalize to other transformations.  
RoIAttn~\cite{liang2022excavating} proposes to enhance the RoI features by passing them through multiple self-attention layers. 
However, simply doing so is limited w.r.t. the ability to obtain invariant RoI features for two reasons: 
1) performing self-attention on RoI Align extracted features has more limited flexibility than sampling on the original feature map,
2) the regression ability of typical self-attention is insufficient for identifying specific semantic regions under different transformations.
Our SRA obtains different semantic regions with a novel semantic attention structure by leveraging the global and local semantic relationship within the RoI. 
We then sample the RoI features from the semantic regions, 
which makes it easy to achieve invariance under more diverse transformations, and thus obtain higher detection accuracy than existing methods.

\subsection{Attention Mechanism}
In computer vision, attention can be regarded as an adaptive process, which mimics the human visual system's ability to focus on important regions. 
RAM~\cite{mnih2014recurrent} is the pioneering work to introduce the attention concept in computer vision. 
After that, there have been some works~\cite{hu2018squeeze, zhao2020exploring} exploring the use of attention mechanisms for different computer vision tasks.
Recently, transformer networks, which have achieved great success in natural language processing~\cite{vaswani2017attention}, are explored in computer vision and have shown great potential. 
ViT~\cite{ViT2020} is the first work to bring transformer into computer vision by regarding a 16$\times$16 pixel region as a word and an image as a sentence.
Due to the strong modeling capability of visual transformer networks, they have been applied to various vision tasks
such as image recognition~\cite{liu2021swin, guo2022visual},
object detection~\cite{detr2020,yang2021sampling}, \etc.
In this paper, we introduce a novel semantic attention mechanism to capture invariance RoI features under more variety of transformations.
\section{Methodology}

In this section, we will first introduce the general architecture of the proposed Semantic RoI Align (SRA).
\ygy{
Next, we will detail how the semantic masks of SRA are obtained. 
}
We will then present a dynamic feature sampling method to dynamically sample features for SRA according to different RoI aspect ratios, which improves model accuracy and efficiency.
Finally, the proposed Area Embedding is introduced to replace the previous position embedding, so as to provide more accurate position information for the model.

\subsection{Semantic RoI Align} 

\begin{figure}[!t]
	\centering
	\includegraphics[width=1\linewidth]{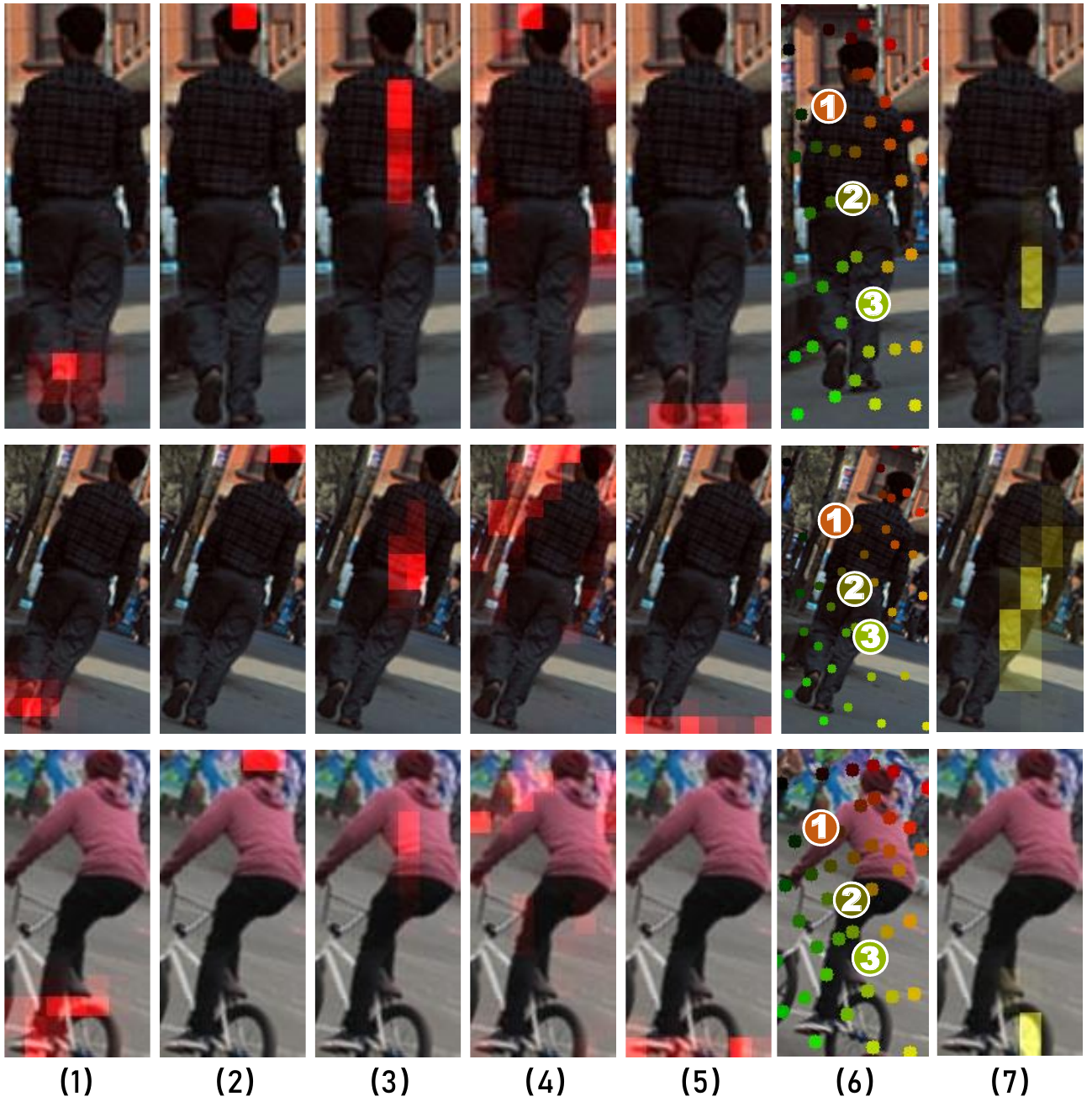}
	\caption{
        Semantic sampling masks of SRA (columns 1 to 5), sampling locations of DRoIPooling~\cite{zhu2019deformable} (column 6), and sampling mask of directly passing features extracted by RoI Align through a standard self-attention layer (column 7). Each row represents an RoI in an image. 
        The object in the middle row is rotated by 30 degrees from the top row, and the bottom row shows another object of the same class as the top row. 
        Red and yellow sampling masks are overlaid on images and the sampling locations of DRoIPooling are indicated with different colors.
    }
	\label{sampling}
\end{figure}

\ygy{
The pipeline of the proposed Semantic RoI Align (SRA) is shown in Figure~\ref{URA}.
The SRA extracted RoI feature of an object consists of N sub-features, each of which is sampled in a specific semantic region, making the sampling position adaptive to image transformations, and thereby improving the transformation-invariance.
In Figure~\ref{sampling}, we visualize partial semantic masks (left 5 columns) produced by our SRA for 3 RoI proposals (one for each row).
The semantic samplings of SRA can sample on the same semantic parts for the object under different perspective transformations such as rotation (top row and middle row) and object pose transformations (top row and bottom row), giving the extracted RoI feature better transformation-invariance, and thus being beneficial to bounding boxes regression and semantic labels prediction in the second-stage network.}

\ygy{
The inputs of SRA are a feature map $F$ with shape $(C, H, W)$ where $C$, $H$ and $W$ represent the number of feature channels, height, and width of the feature map, respectively, and a list of RoI proposals $R=\{R_i\}$ where $R_i=\{x_{i,0}, y_{i,0}, x_{i,1}, y_{i,1}\}$ indicates a bounding box in the feature map with $(x_{i,0}, y_{i,0})$ and $(x_{i,1}, y_{i,1})$ being the coordinates of the top left corner and the bottom right corner, respectively.
For each RoI proposal $R_i$, SRA first exploits the Dynamic Feature Sampler to sample a feature map $f_i$ from the input feature map $F$ with the bounding box $R_i$.
We then obtain $N$ semantic masks $m_i=\{m_{i, n}\}$, $1\leq n \leq N$,
which have the same size as $f_i$. 
The output transformation-invariant RoI feature $y_i$ of our SRA  is finally obtained by sampling $f_i$ using the semantic masks $m_i$. More specifically, $y_i$ is calculated as the weighted sum of $f_i$ elements using $m_i$ elements as weights:
\begin{equation}
\begin{aligned}
&
y_{i}(n,c)=\sum_{j=1}^{h_i} \sum_{k=1}^{w_i} f_i(c,j,k) \cdot  m_{i,n}(j,k),
\\
&\text{for all }n \in \{1,...,N\}, c \in \{1,...,C\},
\end{aligned}
\label{sample2}
\end{equation}
where $h_i, w_i$ are the height and width of feature map $f_i$.
Next, we will introduce how the semantic masks $m_i$ are estimated in the SRA.}

\ygy{
\subsection{Obtaining Semantic Masks}}
\ygy{
The pipeline of obtaining semantic masks of SRA is also shown in Figure~\ref{URA}.
The goal of SRA is to generate $N$ separable semantic part masks $m_i$ for the input RoI proposal $R_i$ and the sampled feature map $f_i$. 
To achieve this, we want the value of $m_{i,n}(j, k)$ to be positively correlated with the likelihood that position $(j,k)$ in $R_i$ belongs to the $n$-th semantic part of the object in $R_i$. Let us denote that likelihood as $m_{i,n}'(j, k)$. The likelihood is related to two factors, namely, 1) what it is in $R_i$, and 2) what it is at the position $(j,k)$ of $R_i$.
The former is expressed 
by a $K$-dimensional RoI descriptor $d_i$ representing the overall features of $R_i$, and the latter is characterized by a semantic feature map $s_i$ with shape $(K, h_i, w_i)$ meaning the semantic feature at different positions in $R_i$.
To make the final RoI feature computed based on Eq.~\ref{sample2} transformation-invariant, the $m_i$ should transform accordingly when the object transforms.
Therefore, we obtain the likelihood $m_i'$ by using a same regressor at different positions $(j,k)$:
\begin{equation}
\begin{aligned}
&m_{i,n}'(j, k)=\xi_n([d_i,s_i(:,j,k)]),\\
&\text{for all }n \in \{1,...,N\}, j \in \{1,...,h_i\}, k \in \{1,...,w_i\},
\label{mask_reg}
\end{aligned}
\end{equation}
where the $\xi$ are $N$ learnable sub-mask regressors, each composed of two lots of Norm-ReLU-Linear, $[\cdot,\cdot]$ means the concatenate operation, and $s_i(:,j,k)$ represents the semantic feature in the position $(j,k)$.
By doing so, if some transformation of the object causes the feature at position $(j,k)$ to move to position $(j',k')$, the transformed masks $\hat{m}_{i,n}(j',k')=\xi_n([\hat{d}_i,\hat{s}_i(:,j',k')])$ will have similar value with $m_{i,n}(j, k)$, since the transformed $\hat{d_i}$ and $\hat{s_i}(:,j',k')$ are similar to  $d_i$ and $s_i(:,j,k)$, respectively. 
This means the semantic masks transform accordingly with the transformation.
We obtained $s_i$ by performing a 1$\times$1 convolution on $f_i$, and we explored various forms of the RoI Descriptor Regressor to obtain $d_i$:
\begin{equation}
\begin{aligned}
&\text{- Concatenation: }d_i=\psi (\flatten (f_i))\\
&\text{- Maximum: }d_i=\psi (\max^{h_i}_{j=1}\max^{w_i}_{k=1} f_{i,(:,j,k)})\\
&\text{- Average: }d_i=\psi (\frac{1}{h_i\times w_i}\sum^{h_i}_{j=1}\sum^{w_i}_{k=1} f_{i,(:,j,k)})
\end{aligned}
\label{RDR}
\end{equation}
Here $\psi$ is a linear layer with $K$ output channels.}

Sampling features based on semantic masks may cause the model to lose position information, which is important for the object detection task. We thus use a position embedding $p_i$ (see \cite{zhao2020exploring}) to provide position information for the model, and use positional embedded $m_{i,n}'(j, k)=\xi_n([d_i,s_i(:,j,k),p_i(:,j,k)])$ instead of Eq.~\ref{mask_reg}. The $p_i$ is obtained by performing a 1$\times$1 convolution with output channels of $P$ on $p'_i$, where $p'_i$ with shape $(2, h_i, w_i)$ is the relative position of each position in the RoI, and is normalized to $[-1, 1]$:
\begin{equation}
\begin{aligned}
p'_i(1,j,k)=&j/h_i\times 2-1,\\
p'_i(2,j,k)=&k/w_i\times 2-1,
\end{aligned}
\label{pos_enc}
\end{equation}
The semantic masks $m_i$ is then obtained by $m_i=softmax(m_i'\cdot\gamma)$, where $\gamma$ is an amplification factor that amplifies the backpropagation response of masks, and the softmax acts on the last two dimensions to ensure the sum of each mask to 1.
Finally, the output RoI feature $y_i$ is obtained by summing up elements of $f_i$ weighted by the $N$ semantic masks $m_i$ as shown in Eq.~\ref{sample2}.

\begin{figure}[!t]
	\centering
	\includegraphics[width=0.9\linewidth]{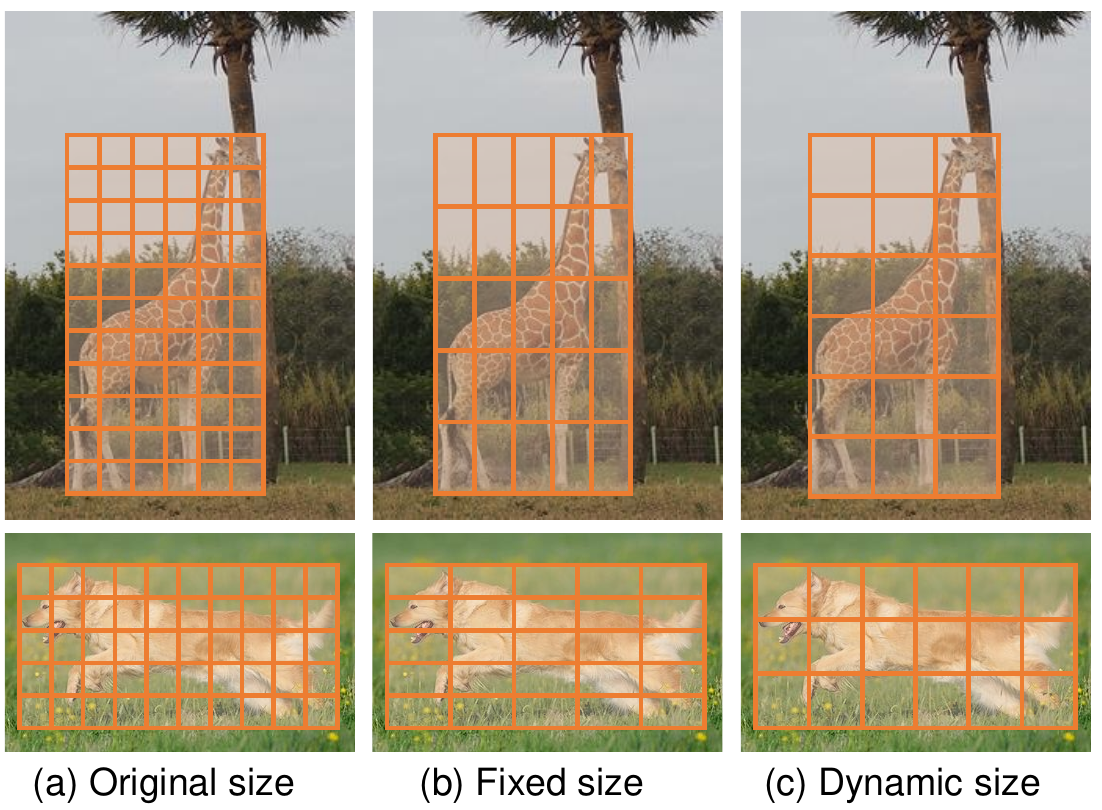}
	\caption{ 
 Different methods to determine the size of the sub-sampled feature map.
 (a) Using a size that is the same resolution as the original feature map is costly as it results in too many samples. 
 (b) Using a fixed size may cause the aspect ratio of the region represented by each sub-sampled feature to be inconsistent for different RoIs, which we believe is harmful to the model, e.g. a ratio of approximately 2 for the upper RoI and 0.5 for the lower RoI.
 (c) Our Dynamic Feature Sampler overcomes the limitations of the above two methods, yielding both consistent and limited samples.}
	\label{dynamic}
\end{figure}

\subsection{Dynamic Feature Sampler}
In SRA, the semantic masks $m_i$ are estimated via the sub-sampled feature map $f_i$, and thus the computational overhead of SRA is proportional to the input size of $f_i$, i.e. $h_i\times w_i$.
The size of feature map $f_i$ can be set to different values for different RoIs.
A straightforward solution is to set the size 
to the original resolution of $R_i$.
However, this would lead to a large computational cost for some large RoIs.
Another way is to set the size to a fixed value. However, as shown in Figure~\ref{dynamic}(b), 
this may cause the aspect ratio of the region represented by each feature of $f_i$ to be inconsistent for different RoIs, 
and experiments show that this will lead to loss of accuracy. 
To balance sampling quality and computational efficiency, we propose a Dynamic Feature Sampler to select the size of $f_i$ for each proposal, which keeps the aspect ratio of each sub-sampled region close to 1, and has a limited size.
Specifically, for each RoI $R_i$, we pick the size that has the closest aspect ratio to $R_i$ while not exceeding a maximal area $M$. Mathematically, this can be formulated as:
\begin{equation}
h_i, w_i = \argmin_{\substack{(h', w') \in \mathbb{Z_+}^2\\ \text{with }  h'\cdot w' \leq M} }\abs{\frac{h'}{w'} - \frac{x_{i,1}-x_{i,0}}{y_{i,1}-y_{i,0}}}.
\end{equation}
The sub-sampled feature map $f_i$ is then obtained by dividing $R_i$ region of $F$ into $h_i\times w_i$ blocks and averaging the feature values in each block.
With the Dynamic Feature Sampler, our SRA yields a good performance with a small computational overhead.

\begin{figure}[!t]
	\centering
	\includegraphics[width=1\linewidth]{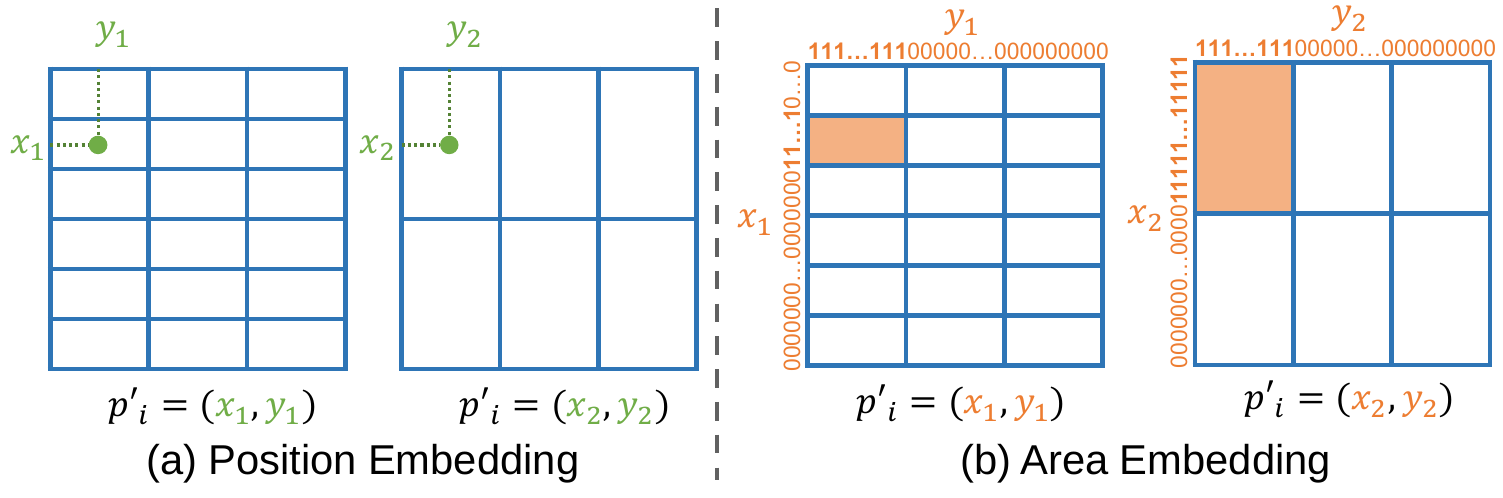}
	\caption{Schematic diagram comparison of traditional Position Embedding and proposed Area Embedding.  (a) Position embedding only embeds the coordinates of the sampling center, while the same center position may represent different sampling areas. (b) Our Area Embedding embeds the entire sampling area. }
	\label{area}
\end{figure}

\subsection{Area Embedding}
In Eq.~\ref{pos_enc}, we use position embedding of each grid center in the mask to provide sampling position information to the model.
However, as shown in Figure~\ref{area}(a), since we use a dynamic way to determine the size of the sub-sampled feature map, the same center position may represent different sampling areas. We thus propose Area Embedding to encode the sampled area of each point in the output feature with two fixed-length vectors, each representing both the position and the coverage on the horizontal and vertical axes. 
We set the length of this vector to $M$, which is the maximal number of samples per axis.
For each point $(j, k)\in \mathbb{Z}^2$ sampled by SRA, we calculate $p^\prime_i$ by: 
\begin{equation}
\begin{aligned}
p'_i(1\cdots M,j,k)= & \upsample\left(\onehot(j; h_i); M\right),\\
p'_i((M+1)\cdots 2M,j,k)=& \upsample\left(\onehot(k; w_i); M\right).
\end{aligned}
\label{area_enc}
\end{equation}
where the $\onehot(b; a)$ operator takes an integer $b$ less or equal to $a$ as input and produces the one hot embedding of $b$ within a vector of length $a$, and $\upsample(\mathbf{v}; M)$ upsamples vector $\mathbf{v}$ to a $M$-sized vector. 
The upsampling method can vary; in Figure~\ref{area}, we use nearest sampling for convenience of illustration, while in our experiments we use linear sampling for higher accuracy. 
The Area Embedding provides the model more accurate sampling position information and experiments show that it improves the accuracy of the model.
\section{Experiments}
We conduct our experiments on the MS COCO dataset~\cite{lin2014microsoft},
and use the train2017 for training and use the val2017 and the test2017 for testing.
We report the standard COCO evaluation metrics including mean Average Precision (AP) under different Intersection over Union (IoU) thresholds and at different object scales, denoted as $AP$ for the object detection task and $AP^{m}$ for the instance segmentation task. Our model is implemented based on Jittor~\cite{hu2020jittor} and JDet\footnote{https://github.com/Jittor/JDet}.
The implementation details of our model are given in the supplementary material.

\subsection{Ablation Studies}
We first conduct a series of ablation experiments to verify the effectiveness of each part of the proposed model. The ablation experiments are conducted on the object detection task using the MS COCO validation set. We couple our proposed feature extractor with Faster-RCNN using ResNet-50 as the backbone. RoI Align~\cite{he2017mask} is used as our baseline model, if not specifically mentioned.

\begin{table}[!t]\centering
	\renewcommand \arraystretch{1.3}
	\renewcommand\tabcolsep{3pt}
	\begin{center}
		\caption{ Results of comparative and ablation experiments between our SRA and baseline models. }
		\label{overall}
	    \scalebox{0.8}{
		\begin{tabular}{l|lll|lll|ll}
			\hline
			    Method & $\AP$ & $\AP_{50}$ & $\AP_{75}$ & $\AP_S$ & $\AP_{M}$ & $\AP_{L}$ & Params & FLOPs \\
			\hline
        		RoI Align & 37.5 & 58.2 & 40.8 & 21.8 & 41.1 & 48.1 & 41.8M & 340.9G\\
        		DRoIPooling & 37.9 & 59.4 & 41.8 & 22.4 & 41.4 & 49.3 & 149.4M & 349.0G\\
                w/ Conv. & 36.4 & 57.4 & 39.6 & 21.5 & 40.0 & 46.9 & 71.9M & 350.1G\\
                w/ SA & 37.6 & 58.0 & 40.8 & 20.9 & 41.1 & 48.7 & 42.7M & 348.1G\\
        		SRA (Ours) & \textbf{39.2} & \textbf{59.6} & \textbf{42.6} & \textbf{22.5} & \textbf{42.6} & \textbf{51.9} & 42.0M & 344.2G\\
			\hline
		\end{tabular}
		}
	\end{center}
\end{table}

\textbf{The effectiveness of SRA.} To verify the effectiveness of the proposed SRA, we replaced it with RoI Align and Modulated Deformable RoI Pooling (DRoIPooling)~\cite{zhu2019deformable}. 
The results in Table~\ref{overall} show that, our model outperforms the baseline model by 1.7\% AP with a minor computational and parameters cost, and also outperforms the DRoIPooling by 1.3\% AP with a much smaller model.

We also compared SRA with two other baselines, by performing some simple operations on the features extracted by RoI Align: 
applying a convolutional layer on the features to obtain the sampling masks and re-sampling the features with these masks 
(denoted as ``w/ Conv.''), as well as 
directly passing the features through a standard multi-head self-attention layer~\cite{vaswani2017attention} (denoted as ``w/ SA''). 
The results in Table~\ref{overall} show that our model achieves a gain of 2.8\% and 1.6\% in AP, respectively, with a smaller number of parameters and FLOPs.
This improvement can be attributed to its enhanced capability in identifying consistent semantic parts across diverse transformations, leading to better transformation-invariance.

We also visualize some samplings of SRA, DRoIPooling, and ``w/ SA'' in Figure~\ref{sampling}.
Rows 1 \& 2 together show how samplings respond to different transformations of the same object, while rows 1 \& 3 together indicate how samplings respond to different objects of the same class.
We found the sampling masks of SRA (columns 1 to 5) can be divided into two classes. The first class samples on different semantic parts. For example, columns 1-4 show the samples on the human's feet, head, and body, and around the human, respectively. 
The second class of sampling is for positioning, which is only activated in certain positions. For example, the 5th column is only activated on the bottom position of the RoI. 
Our semantic samplings can sample on the same semantic parts for the object under different transformations, which gives the extracted RoI feature better transformation invariance. 
In comparison, the sampling locations of DRoIPooling (column 6 of Figure~\ref{sampling}) are distributed mostly inside the object as the object transforms, however, they will not vary accordingly with object pose transformations. Taking 3 samplings (in column 6) as an example, in the top row and middle row, the circles numbered from 1 to 3 do not rotate with the object, which means DRoIPooling can not always achieve transformation-invariance under some complex transformations like rotation and object pose changes.
Also, simply passing features through a self-attention layer (w/ SA, column 7 of Figure~\ref{sampling}) cannot ensure sampling on the same semantic parts, thus failing to obtain transformation-invariant RoI features.

\begin{table}[!t]\centering
	\renewcommand \arraystretch{1.3}
	\renewcommand\tabcolsep{7pt}
	\begin{center}
		\caption{ Ablation study on the effectiveness of each module in our SRA. D, S, and A denote RoI descriptor, semantic feature map, and Area Embedding respectively.}
		\label{components}
	    \scalebox{0.8}{
		\begin{tabular}{lll|lll|lll}
			\hline
			    D & S & A & $\AP$ & $\AP_{50}$ & $\AP_{75}$ & $\AP_S$ & $\AP_{M}$ & $\AP_{L}$ \\
			\hline
                \cmark & \xmark & \xmark & 31.4 & 52.2 & 32.7 & 17.8 & 35.1 & 40.0\\
                \xmark & \cmark & \xmark & 36.2 & 57.5 & 38.6 & 21.0 & 39.8 & 46.6\\
                \xmark & \xmark & \cmark & 37.3 & 58.3 & 40.7 & 21.5 & 41.0 & 48.1\\
			\hline
                \xmark & \cmark & \cmark & 38.9 & 59.3 & 42.3 & 22.5 & 42.5 & 51.3\\
                \cmark & \xmark & \cmark & 37.4 & 58.4 & 40.4 & 22.1 & 41.0 & 48.4\\
                \cmark & \cmark & \xmark & 36.4 & 57.3 & 39.0 & 20.9 & 39.9 & 46.7\\
                \cmark & \cmark & PE & 38.8 & 59.2 & 42.5 & \textbf{22.6} & 42.3 & 50.7\\
			\hline
                \cmark & \cmark & \cmark & \textbf{39.2} & \textbf{59.6} & \textbf{42.6} & 22.5 & \textbf{42.6} & \textbf{51.9}\\
			\hline
		\end{tabular}
		}
	\end{center}
\end{table}

\begin{table}[!t]\centering
	\renewcommand \arraystretch{1.3}
	\renewcommand\tabcolsep{8pt}
	\begin{center}
		\caption{ Experiments on different module settings. DR denotes RoI Descriptor Regressor, and $\gamma$ denotes the amplification factor.  }
		\label{gamma}
	    \scalebox{0.8}{
		\begin{tabular}{l|lll|lll}
			\hline
			    Setting & $\AP$ & $\AP_{50}$ & $\AP_{75}$ & $\AP_S$ & $\AP_{M}$ & $\AP_{L}$  \\
			\hline
                DR=Con. & 38.9 & 59.3 & 41.9 & \textbf{22.5} & 42.3 & 51.1\\
                DR=Max. & 39.0 & 59.3 & \textbf{42.9} & \textbf{22.5} & \textbf{42.6} & 51.1\\
                DR=Avg. & \textbf{39.2} & \textbf{59.6} & 42.6 & \textbf{22.5} & \textbf{42.6} & \textbf{51.9}\\
			\hline
                $\gamma=1$ & 38.3 & 58.7 & 41.8 & 22.2 & 41.8 & 50.2\\
                $\gamma=5$ & 38.7 & 59.2 & 42.0 & 22.1 & 42.5 & 51.0\\
                $\gamma=50$ & \textbf{39.2} & \textbf{59.6} & \textbf{42.6} & \textbf{22.5} & \textbf{42.6} & \textbf{51.9}\\
                $\gamma=500$ & 36.8 & 57.6 & 39.9 & 21.2 & 40.2 & 48.1\\
			\hline
		\end{tabular}
		}
	\end{center}
\end{table}

\begin{table}[!t]\centering
	\renewcommand \arraystretch{1.3}
	\renewcommand\tabcolsep{2.8pt}
	\begin{center}
		\caption{ 
  Comparison between SRA with a different number of masks (N) and the baseline model with comparable RoI sizes. }
		\label{N}
	    \scalebox{0.8}{
		\begin{tabular}{l|lll|lll|ll}
			\hline
			    Setting & $\AP$ & $\AP_{50}$ & $\AP_{75}$ & $\AP_S$ & $\AP_{M}$ & $\AP_{L}$ & Params & FLOPs \\
			\hline
                $N=9$ & 38.0 & 58.2 & 41.3 & 21.8 & 41.3 & 49.9 & 31.5M & 340.7G\\
                $N=25$ & 38.6 & 59.0 & 41.9 & 22.8 & 42.3 & 50.5 & 35.7M & 342.1G\\
                $N=49$ & 39.2 & 59.6 & 42.6 & 22.5 & \textbf{42.6} & \textbf{51.9} & 42.0M & 344.2G\\
                $N=100$ & \textbf{39.4} & \textbf{59.9} & \textbf{42.9} & \textbf{22.9} & \textbf{42.6} & 51.7 & 55.4M & 348.7G\\
			\hline
                $size=3\times 3$ & 36.3 & 57.3 & 39.0 & 20.6 & 39.9 & 46.4 & 31.3M & 337.8G\\
                $size=5\times 5$ & 37.2 & 58.0 & 40.7 & 21.4 & 40.8 & 47.8 & 35.5M & 339.0G\\
                $size=7\times 7$ & 37.3 & 58.3 & 40.7 & 21.5 & 41.0 & 48.1 & 41.8M & 340.9G\\
                $size=10\times 10$ & 37.6 & 58.4 & 40.8 & 22.0 & 41.3 & 48.8 & 55.1M & 344.9G\\
			\hline
		\end{tabular}
		}
	\end{center}
\end{table}

\ygy{
\textbf{Structure of SRA and Area Embedding.}
We also conduct experiments to verify the effectiveness of different components in the SRA by controlling whether to concatenate the RoI descriptor (D), semantic feature map (S), and Area Embedding (A) when regressing the masks. The results are listed in Table~\ref{components}. }
Comparing the last row with the 5th row in the table, our model with semantic feature map obtains a gain of 1.8\% in AP, as our model determines the masks for sampling based on semantic features, which makes the sampled features invariant under a variety of transformations, thus achieving better performance.
We also tested our model with or without Area Embedding (8th row and 6th row respectively) and replaced the Area Embedding with position embedding (denoted as PE, 7th row).
The results show that the model with AE obtains 2.8\% higher AP than without AE, and 0.4\% higher AP than with PE, which demonstrates AE can describe more accurately the sampling information of Dynamic Feature Sampler and provide better position information for the model.

\begin{table}[!t]\centering
	\renewcommand \arraystretch{1.3}
	\renewcommand\tabcolsep{6.5pt}
	\begin{center}
		\caption{ Experiments on the Dynamic Feature Sampler. M denotes the dynamic feature map size limit. }
		\label{DFS}
	    \scalebox{0.8}{
		\begin{tabular}{l|lll|l|ll}
			\hline
			    Setting & $\AP$ & $\AP_{50}$ & $\AP_{75}$ & Avg. S & Params & FLOPs \\
			\hline
                fixed & 38.8 & 59.2 & 42.4 & 64 & 42.0M & 344.1G\\
                $M=32$ & 38.3 & 59.1 & 42.0 & 16 & 42.0M & 341.7G\\
                $M=64$ & 38.8 & 59.5 & 42.5 & 32 & 42.0M & 342.5G\\
                $M=128$ & \textbf{39.2} & \textbf{59.6} & \textbf{42.6} & 64 & 42.0M & 344.2G\\
                $M=256$ & 39.1 & 59.5 & 42.5 & 128 & 42.0M & 347.9G\\
			\hline
		\end{tabular}
		}
	\end{center}
\end{table}

\begin{table}[!t]\centering
	\renewcommand \arraystretch{1.3}
	\renewcommand\tabcolsep{6.5pt}
	\begin{center}
		\caption{ Comparison with different RoI extractors on the MS COCO detection test-dev set. }
		\label{roi-ext}
	    \scalebox{0.8}{
		\begin{tabular}{l|lll|lll}
			\hline
			    Method & $\AP$ & $\AP_{50}$ & $\AP_{75}$ & $\AP_S$ & $\AP_{M}$ & $\AP_{L}$   \\
			\hline
        		RoI Pooling & 37.2 & 58.9 & 40.3 & 21.5 & 40.2 & 46.0 \\
        		RoI Align & 37.7 & 58.9 & 40.6 & 21.9 & 40.7 &  46.4 \\
        		DRoIPooling & 38.1 & 60.0 & 42.0 & 22.0 & 41.2 & 47.2 \\
        		Ada. RoI Align & 37.7 & 58.8 & 40.7 & 21.8 & 40.7 & 46.5 \\
        		Pr. RoI Align & 37.8 & 58.9 & 40.9 & 22.1 & 40.8 & 46.7 \\
                RoIAttn & 38.0 & 59.3 & 40.9 & 22.4 & 41.1 & 46.9\\
        		SRA & \textbf{39.2} & \textbf{59.8} & \textbf{42.6} & \textbf{22.6} & \textbf{42.1} & \textbf{49.0} \\
			\hline
		\end{tabular}
		}
	\end{center}
\end{table}

\begin{table}[!t]\centering
	\renewcommand \arraystretch{1.3}
	\renewcommand\tabcolsep{8pt}
	\begin{center}
		\caption{ Comparison between RoI Align (RA) and SRA with different methods and backbones on the MS COCO detection test-dev set. }
		\label{sota-det}
	    \scalebox{0.8}{
		\begin{tabular}{c|c|c|c|c}
			\hline
                \multirow{2}{*}{Method} & \multirow{2}{*}{Backbone} & \multirow{2}{*}{Iterations} 
                & \multicolumn{2}{c}{$\AP$} \\
                \cline{4-5}
                &&& RA & \textbf{SRA} \\
			\hline
                \multirow{4}{*}{FRCNN} & R-50 & 1x & 37.7 & 39.2$_{(+1.5)}$ \\
                 & R-101 & 1x & 39.7 & 41.2$_{(+1.5)}$ \\
                 & RX-101-32 & 1x & 41.3 & 42.5$_{(+1.2)}$ \\
                 & RX-101-64 & 1x & 42.6 & 43.7$_{(+1.1)}$ \\
			\hline
                \multirow{5}{*}{MRCNN} & Swin-T & 3x & 46.3 & 47.3$_{(+1.0)}$ \\
                 & Swin-S & 3x & 48.9 & 49.4$_{(+0.5)}$ \\
                 & Swin-B & 3x & 49.3 & 49.7$_{(+0.4)}$ \\
                 & PVTv2-B0 & 1x & 38.4 & 40.0$_{(+1.6)}$ \\
                 & ViT-Adap.-T & 1x & 41.5 & 42.4$_{(+0.9)}$ \\
                 & InternImg.-T & 1x & 47.4 & 48.3$_{(+0.9)}$ \\
			\hline
                DyHead & R-50 & 1x & 40.7 & 41.6$_{(+0.9)}$ \\
			\hline
		\end{tabular}
		}
	\end{center}
\end{table}

\textbf{Choices of the RoI Descriptor Regressor.} We tested various choices of the RoI Descriptor Regressor in Eq.~\ref{RDR}, denoted as DR=Con., Max., and Avg., respectively, where the choice of \emph{concatenation} is tested on the $8\times8$ fixed size feature sampler as it cannot be adapted to the Dynamic Feature Sampler. The results are shown in Table~\ref{gamma}. 
Though the choice of DR=Con. shows slightly better performance than $8\times8$ fixed size feature sampler with the choice of \emph{average} (38.8\% AP in Table~\ref{DFS}), considering that it cannot be adapted to the Dynamic Feature Sampler and will lead to a larger amount of calculation and more parameters, we finally use the \emph{average} RoI Descriptor Regressor in our model.

\textbf{Parameters setting.} The number of masks $N$ determines the size of the RoI features extracted by our model. We tested different settings of $N$ (denoted as ``$N=\text{x}$'') and compared them with the baseline model with different settings of RoI Align output size  (denoted as ``$size=\text{x}$''). The results in Table~\ref{N} show that, with the same RoI feature size (comparing the 1st row with the 5th row, the 2nd row with the 6th row, etc.), our model has a 1.4\%-1.9\% higher AP, 
which proves that the transformation-invariant features extracted by our model contain richer information under the same feature length and are more conducive to object detection. 
Considering the balance between model parameters and accuracy, we finally set $N=49$.
We also tested different settings of the amplification factor $\gamma$, denoted as ``$\gamma=\text{x}$'' in Table~\ref{gamma}. Results show that setting it to an appropriate value is beneficial to the regression of the semantic mask, so we set $\gamma=50$ according to the experiment.

\textbf{Dynamic Feature Sampler.} To evaluate the effectiveness of the Dynamic Feature Sampler, we compared $8\times8$ fixed size feature sampler (denoted as fixed) with Dynamic Feature Sampler ($M=128$), which has the same number of average samplings and similar FLOPs. As shown in the 1st and 4th row of Table~\ref{DFS}, the Dynamic Feature Sampler obtained better results as its sub-sampled feature represented region has a more consistent aspect ratio. 
We also tested different dynamic feature map size limit $M$. 
A larger $M$ brings a higher feature sampling resolution. In general, the accuracy increases with the increment in the resolution; however, the accuracy improvement brought by the increase in sampling resolution is limited by the resolution of the original feature map, and will gradually tend to zero.
So, we choose $M=128$ based on the experiment.

\subsection{Comparison with Other Methods}
We also compared our model with other methods on the COCO test set. We first compared ours with different RoI feature extractors: RoI Pooling~\cite{girshick2015fast}, RoI Align~\cite{he2017mask}, Adaptive RoI Align~\cite{jung2018real}, Precise RoI Align~\cite{jiang2018acquisition}, DRoIPooling~\cite{zhu2019deformable}, and RoIAttn~\cite{liang2022excavating} on Faster R-CNN with ResNet50 as backbone, trained for 12 epochs. The results are shown in Table~\ref{roi-ext}. 
One can see that our model achieves the best performance. In particular, compared to the RoIAttn which incorporates several self-attention layers on the RoI Align extracted feature, our model obtains 1.2\% higher AP. The results demonstrate the advantage of our SRA method for extracting RoI features that are invariant to more types of transformations. 

To verify the generalizability of our method, we also examined the performance gain by using SRA across different detection methods and backbones. The detection methods considered include Faster R-CNN~\cite{ren2015faster}, Mask R-CNN~\cite{he2017mask} and Dynamic Head~\cite{dai2021dynamic}, denoted as FRCNN, MRCNN, and DyHead, respectively. The different backbones considered are: ResNet~\cite{he2016deep}, ResNeXt~\cite{xie2017aggregated}, Swin~\cite{liu2021swin}, PVTv2~\cite{wang2021pvtv2}, ViT-Adapter~\cite{chen2022vitadapter}, and InternImage~\cite{wang2022internimage}, denoted as R, RX, Swin, PVTv2, ViT-Adap., and InternImg., respectively. The setting of hyper-parameters followed the configuration of baseline models. Some $\AP$ results are shown in Table~\ref{sota-det}. Please refer to the supplementary material for more results. The results show that our model improves the accuracy of various detection methods and backbone networks, demonstrating the generalizability of our model.
\section{Conclusions}
In this paper, we proposed SRA, a transformation-invariant RoI feature extractor. It regresses semantic masks based on a novel semantic attention structure, and obtains RoI features by sampling the feature map with these semantic masks, making it invariant under more diverse transformations. We further proposed the Dynamic Feature Sampler to speed up the process while minimizing the impact on accuracy, and proposed Area Embedding to provide more accurate sampling area information. 
Benefiting from the capability and generalizability of SRA, experiments show that its utilization can bring significant performance improvement to various baselines and state-of-the-art models with a small computational and parameter overhead.
\section{Acknowledgments}
This work was supported by the National Key Research and Development Program of China under Grant 2021ZD0112902, the National Natural Science Foundation of China under Grant 62220106003, and the Research Grant of Beijing Higher Institution Engineering Research Center and Tsinghua-Tencent Joint Laboratory for Internet Innovation Technology.
\section{Appendix}

\subsection{Implementation details}
The proposed Semantic RoI Align (SRA) can be used as a plugin to replace the RoI feature extractor of most two-stage object detection networks. In the object detection task, we use Faster R-CNN with different backbones as the baseline models. 
In the instance segmentation task, we use Mask R-CNN with different backbones as the baselines, and  only replace the RoI feature extractor of the object detection head with the SRA.
The computational cost in the experiments counts the floating-point operations per second (FLOPs) of networks with 300 candidate RoIs.
We set the number of masks $N=49$ which makes the input of the second stage network the same size as other RoI feature extractors for a fair comparison.
We set the dynamic feature map size limit $M=128$, the size of RoI descriptor $K=256$, the amplification
factor $\gamma=50$. 
The channel size $C$ is determined by the channel size of extracted feature map of the baseline model.
Following the baseline models, we trained our model using SGD optimizer with an initial learning rate of 0.02, a momentum of 0.9, and a weight decay of 0.0001. 
When using Swin Transformer~\cite{liu2021swin} as the backbone, we trained our model using AdamW optimizer with an initial learning rate of 0.0001, $\beta_1=0.9$, $\beta_2=0.999$, and a weight decay of 0.05 as its default setting. We trained our model for 12 epochs (1x) by default, and the learning rate is stepped down by 0.1 at 67\% and 92\% of training epochs.
Our settings for other hyper-parameters follow the configuration of baseline models.

\subsection{Visual ablation experiments}
\begin{figure}[!t]
\begin{center}
   \includegraphics[width=1\linewidth]{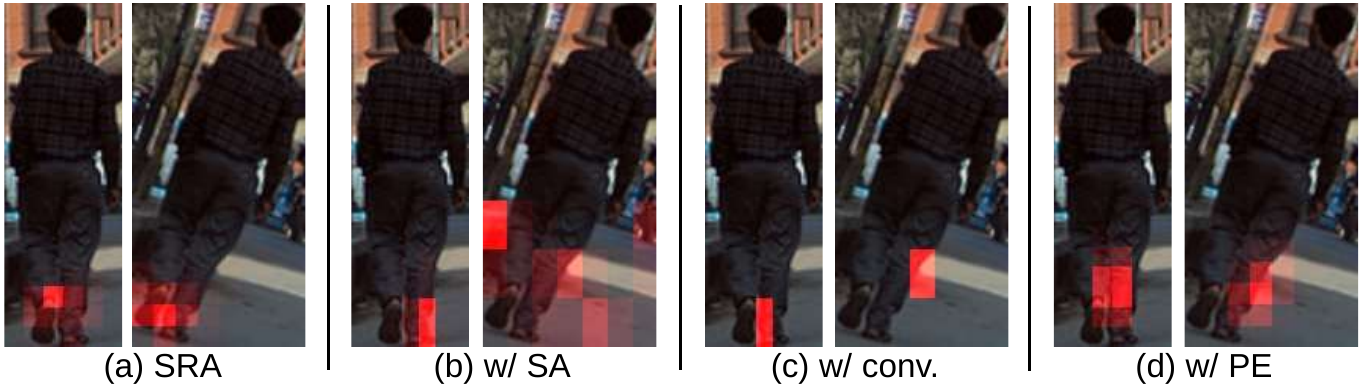}
\end{center}
   \caption{Visual ablation experiments on sampling masks. For each group, the input of the right image is rotated by 30 degrees from that of the left image.
   }
\label{seg}
\end{figure}

In Figure~\ref{seg}, we visualize the sampling masks under rotation transformation of (a) SRA, 
(b) directly passing the features extracted by RoI Align through a standard multi-head self-attention layer \textbf{(w/ SA)}, 
(c) using a convolutional layer to obtain the sampling masks for the RoI Align extracted features \textbf{(w/ conv.)}, 
and (d) replacing Area Embedding of SRA with Position Embedding \textbf{(w/ PE)}. 
The results in (a-c) show that our SRA can have a more consistent response to the same semantic region, thus making the extracted features transformation-invariant. (a, d) show that AE provides more accurate position information, thus obtaining more accurate sampling masks. 

\subsection{Results of generalizability experiment}
In Table~7 of the paper, we showed partial results of the generalizability experiment. In Table~\ref{sota-det-supp} and \ref{sota-seg-supp} of this supplementary material, we show the complete results of that experiment. 

To verify the generalizability of the proposed method, we examined the performance improvement gained by using SRA across different detection methods and backbone networks. The detection methods considered include Faster R-CNN~\cite{ren2015faster}, Mask R-CNN~\cite{he2017mask} and Dynamic Head~\cite{dai2021dynamic}, denoted as FRCNN, MRCNN, and DyHead, respectively. The different backbones considered are ResNet~\cite{he2016deep}, ResNeXt~\cite{xie2017aggregated}, Swin~\cite{liu2021swin}, PVTv2~\cite{wang2021pvtv2}, ViT-Adapter~\cite{chen2022vitadapter}, and InternImage~\cite{wang2022internimage}, denoted as R, RX, Swin, PVTv2, ViT-Adap., and InternImg., respectively. The setting of other hyper-parameters followed the same configuration as baseline models.  The results show that semantic-aware RoI Align is useful and can improve the accuracy of various detection methods and backbone networks, including some attention (DyHead, InternImg., etc.) and self-attention (Swin, PVTv2, etc.) based models, demonstrating the generalizability of our model.

\begin{table*}[!t]\centering
	\renewcommand \arraystretch{1.3}
	\renewcommand\tabcolsep{14pt}
	\begin{center}
		\caption{ Comparison with different methods on the MS COCO detection test-dev set. }
		\label{sota-det-supp}
	    \scalebox{0.8}{
		\begin{tabular}{l|l|l|l|lll|lll}
			\hline
			    Method & RoI Extractor& Backbone & Iterations & $\AP$ & $\AP_{50}$ & $\AP_{75}$ & $\AP_S$ & $\AP_{M}$ & $\AP_{L}$   \\
			\hline
        		FRCNN & RoI Align& R-101 & 1x & 39.7 & 60.7 & 43.2 & 22.5 & 42.9 & 49.9 \\
        		FRCNN&\textbf{SRA} & R-101 & 1x & 41.2$_{(+1.5)}$ & 61.8 & 45.0 & 23.5 & 44.1 & 52.3 \\
        		FRCNN & RoI Align& RX-101-32 & 1x & 41.3 & 62.5 & 45.1 & 24.2 & 44.7 & 51.7 \\
        		FRCNN&\textbf{SRA} & RX-101-32 & 1x & 42.5$_{(+1.2)}$ & 63.4 & 46.5 & 24.8 & 45.8 & 53.5 \\
        		FRCNN & RoI Align& RX-101-64 & 1x & 42.6 & 63.8 & 46.5 & 25.4 & 46.0 & 53.3 \\
        		FRCNN&\textbf{SRA} & RX-101-64 & 1x & 43.7$_{(+1.1)}$ & 64.5 & 47.9 & 25.7 & 46.9 & 55.1 \\
        		DyHead & RoI Align & R-50 & 1x & 40.7 & 60.6 & 44.1 & 23.1 & 43.4 & 50.8 \\
        		DyHead & \textbf{SRA} & R-50 & 1x & 41.6$_{(+0.9)}$ & 61.4 & 45.3 & 23.8 & 44.6 & 51.4\\
			\hline
		\end{tabular}
		}
	\end{center}
\end{table*}

\begin{table*}[!t]\centering
	\renewcommand \arraystretch{1.3}
	\renewcommand\tabcolsep{5pt}
	\begin{center}
		\caption{ Comparison with different methods on the MS COCO detection and instance segmentation test-dev set.  }
		\label{sota-seg-supp}
	    \scalebox{0.8}{
		\begin{tabular}{l|l|l|l|lll|lll|lll|lll}
			\hline
			    Method & RoI Extractor & Backbone & Iterations & $\AP$ & $\AP_{50}$ & $\AP_{75}$ & $\AP_S$ & $\AP_{M}$ & $\AP_{L}$ & $\AP^{m}$ & $\AP^{m}_{50}$ & $\AP^{m}_{75}$ & $\AP^{m}_S$ & $\AP^{m}_{M}$ & $\AP^{m}_{L}$  \\
			\hline
        		MRCNN & RoI Align & Swin-T & 3x & 46.3 & 68.7 & 51.0 & 28.8 & 48.9 & 58.2 & 42.0 & 65.8 & 45.3 & 24.5 & 44.6 & 55.0\\
        		MRCNN & \textbf{SRA} & Swin-T & 3x & 47.3$_{(+1.0)}$ & 69.3 & 52.1 & 29.7 & 49.7 & 60.0 & 42.4 & 66.2 & 46.0 & 25.0 & 44.9 & 55.5\\
        		MRCNN & RoI Align & Swin-S & 3x & 48.9 & 70.7 & 53.9 & 30.7 & 52.0 & 61.7 & 43.9 & 68.0 & 47.5 & 25.9 & 46.9 & 57.4\\
        		MRCNN & \textbf{SRA} & Swin-S & 3x & 49.4$_{(+0.5)}$ & 71.0 & 54.3 & 31.1 & 52.4 & 62.9 & 44.0 & 68.3 & 47.6 & 26.2  & 46.9  & 58.0\\
          
        		MRCNN & RoI Align & Swin-B & 3x & 49.3 & 70.8 & 54.3 & 31.1 & 52.2 & 62.2 & 44.2 & 68.2 & 48.0 & 26.5 & 47.1 & 57.5 \\
        		MRCNN & \textbf{SRA} & Swin-B & 3x & 49.7$_{(+0.4)}$ & 71.1 & 54.6 & 30.9 & 52.5 & 62.9 & 44.3 & 68.4 & 47.9 & 26.4 & 47.1 & 58.0\\
          
        		MRCNN & RoI Align & PVTv2-B0 & 1x         & 38.4 & 60.8 & 41.7 & 22.4 & 40.5 & 48.7 & 36.2 & 58.0 & 38.6 & 19.9 & 38.0 & 47.6 \\
        		MRCNN & \textbf{SRA} & PVTv2-B0 & 1x      & 40.0$_{(+1.6)}$ & 61.9 & 43.4 & 23.2 & 42.1 & 51.2 & 36.9 & 59.1 & 39.5 & 20.5 & 38.8 & 48.8 \\
        		MRCNN & RoI Align & ViT-Adap.-T & 1x      & 41.5 & 63.0 & 45.2 & 24.1 & 43.5 & 53.3 & 37.8 & 60.1 & 40.4 & 20.7 & 39.5 & 50.5 \\
        		MRCNN & \textbf{SRA} & ViT-Adap.-T & 1x   & 42.4$_{(+0.9)}$ & 63.7 & 46.2 & 24.5 & 44.5 & 54.8 & 38.3 & 60.7 & 41.1 & 21.1 & 40.1 & 51.3 \\
        		MRCNN & RoI Align & InternImg.-T & 1x     & 47.4 & 69.4 & 52.1 & 29.1 & 50.5 & 60.0 & 43.0 & 66.8 & 46.5 & 25.4 & 45.6 & 56.3 \\
        		MRCNN & \textbf{SRA} & InternImg.-T & 1x  & 48.3$_{(+0.9)}$ & 70.0 & 53.3 & 29.6 & 51.5 & 61.5 & 43.4 & 67.3 & 47.0 & 25.6 & 46.2 & 57.0 \\
			\hline
		\end{tabular}
		}
	\end{center}
\end{table*}

\ygy{
\subsection{Results on more datasets and tasks}
We further conduct experiments on the Pascal VOC dataset, taking Faster R-CNN+ResNet50 as the baseline, and replace the RoI extractor with ours. Results are in the last column of Table~\ref{vid-det}. 
We also perform the video object detection on the ImageNet VID dataset~\cite{russakovsky2015imagenet}, taking FGFA~\cite{zhu2017flow}+Faster R-CNN+ResNet50 as the baseline. The results are in columns 2--7. 
Our experimental settings and hyperparameters followed the baseline.
All these experiments show our SRA significantly outperforms the other RoI extractors, demonstrating its generalizability.
}

\begin{table}[!t]\centering
	\renewcommand \arraystretch{1.3}
	\renewcommand\tabcolsep{6pt}
	\begin{center}
		\caption{ \ygy{Columns 2--7: Video object detection experiment results on the ImageNet VID validation set. Column 8: Experiment results on the Pascal VOC 2007 test set }}
		\label{vid-det}
	    \scalebox{0.8}{
		\begin{tabular}{l|lll|lll|l}
            \hline
                \multirow{2}{*}{Method} & \multicolumn{6}{c|}{ImageNet VID} & VOC\\
                \cline{2-8}
                & $\AP$ & $\AP_{50}$ & $\AP_{75}$& $\AP_{S}$ & $\AP_{M}$ & $\AP_{L}$ & AP\\
            \hline
                RoI Align  & 47.1 & 74.7 & 52.1 & 5.9 & 22.2 & 53.1 & 80.2\\
                SRA (Ours)  & \textbf{48.0} & \textbf{76.5} & \textbf{53.0} & \textbf{6.7} & \textbf{22.7} & \textbf{54.1}  & \textbf{81.8}\\
            \hline
		\end{tabular}
		}
	\end{center}
\end{table}

\begin{figure}[!t]
\begin{center}
   \includegraphics[width=0.6\linewidth]{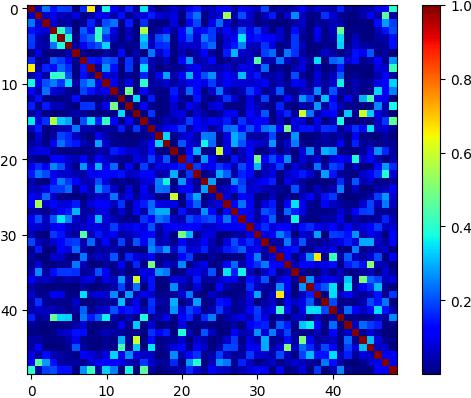}
\end{center}
   \caption{ \ygy{Average cosine similarity between SRA sampling masks on the MS COCO validation set.}
   }
\label{heatmap}
\end{figure}

\ygy{
\subsection{Variety of the sampling masks}
In Figure~\ref{heatmap}, we tested the average cosine similarity between different sampling masks of SRA on the MS COCO validation set, and more than 94\% of the mask-to-mask similarities were less than 0.3, which demonstrates the diversity of SRA masks.
}

\ygy{
\subsection{Invariance of RoI feature}
To test the invariance of RoI features,  we evaluate the cosine similarity between RoI features before and after performing random rotation and horizontal reflection.
We also tested randomly scaling and panning the RoI box, to simulate the error in RoI proposals. Results in Table~\ref{invariance_voc} show that our SRA obtains 17\% -- 32\% better invariance and  the advantage is more obvious for rotation.
}

\begin{table}[!t]\centering
	\renewcommand \arraystretch{1.3}
	\renewcommand\tabcolsep{8pt}
	\begin{center}
		\caption{ Average similarity between RoI features before and after performing different transformations.}
		\label{invariance_voc}
	    \scalebox{0.8}{
		\begin{tabular}{l|lll}
			\hline
			    Method & Rotation & Reflection & Scaling \& Panning  \\
			\hline
                RoI Align  & 0.37 & 0.53 & 0.59 \\
                DRoIPooling  & 0.38 & 0.53 & 0.62\\
                SRA (Ours)  & \textbf{0.49} & \textbf{0.69} & \textbf{0.73}\\
			\hline
		\end{tabular}
		}
	\end{center}
\end{table}

\ygy{
\subsection{Standard deviation experiments}
We conducted the standard deviation experiment in Table~\ref{time} by training and test each model for 4 times.
}

\begin{table}[!t]\centering
	\renewcommand \arraystretch{1.3}
	\renewcommand\tabcolsep{11pt}
	\begin{center}
		\caption{ \ygy{Results of standard deviation experiment.}}
		\label{time}
	    \scalebox{0.8}{
		\begin{tabular}{l|lll}
			\hline
			    Method & $\AP$ & $\AP_{50}$ & $\AP_{75}$ \\
			\hline
                RoI Align  & $37.4 \pm 0.1$ & $58.3 \pm 0.1$ & $40.6 \pm 0.1$\\
                SRA (Ours)  & $\textbf{39.1} \pm 0.1$ & $\textbf{59.4} \pm 0.2$ & $\textbf{42.6} \pm 0.1$\\
			\hline
		\end{tabular}
		}
	\end{center}
\end{table}

\bibliography{aaai24}

\end{document}